\RequirePackage{fix-cm}
\documentclass[twocolumn]{svjour3}          
\smartqed  
\usepackage{graphicx}
\usepackage{natbib}
\usepackage{multirow}
\usepackage{amsmath}
\usepackage{amssymb}
\usepackage{bbm}
\usepackage{bm}
\usepackage{booktabs}
\usepackage{array} 
\usepackage{blindtext}
\usepackage{comment}
\usepackage{makecell}
\usepackage[usestackEOL]{stackengine}

\usepackage{tabularx}
\usepackage{algorithm}
\usepackage{algorithmic}
\usepackage{afterpage}
\usepackage{subcaption}
\usepackage{animate}
\usepackage{xspace}

\makeatletter
\DeclareRobustCommand\onedot{\futurelet\@let@token\@onedot}
\def\@onedot{\ifx\@let@token.\else.\null\fi\xspace}

\def\eg{\emph{e.g}\onedot} 
\def\ie{\emph{i.e}\onedot}

\makeatother

\usepackage{color, colortbl}
\usepackage[table]{xcolor}
\definecolor{remark}{rgb}{1,.5,0} 
\definecolor{citecolor}{rgb}{0,0.443,0.737} 
\definecolor{linkcolor}{rgb}{0.956,0.298,0.235} 
\definecolor{gray}{gray}{0.95}
\definecolor{cyan}{rgb}{0.831,0.901,0.945}
\definecolor{mygray}{gray}{.9}

\definecolor{lightgreen}{HTML}{D8ECD1}
\definecolor{edit}{HTML}{C63678}

\usepackage{pifont}

\usepackage{color, colortbl}
\definecolor{citecolor}{HTML}{0071bc}
\definecolor{tabhighlight}{HTML}{e5e5e5}
\usepackage[colorlinks,citecolor=citecolor]{hyperref}

\makeatletter
\renewcommand\paragraph{
  \@startsection{paragraph} 
  {4} 
  {\z@} 
  {.5em \@plus1ex \@minus.2ex} 
  {-.5em} 
  {\normalfont\normalsize\bfseries} 
}
\makeatother
\begin{document}
\sloppy

\title{Show-1: Marrying Pixel and Latent Diffusion Models for Text-to-Video Generation 
}


\author{
David Junhao Zhang\thanks{hihihi} \and
Jay Zhangjie Wu \and
Jia-Wei Liu \and
Rui Zhao \and
Lingmin Ran \and
Yuchao Gu \and
Difei Gao \and
Mike Zheng Shou
}
\vspace{-100mm}

\date{Received: date / Accepted: date}


\twocolumn[
{%
\maketitle
\setlength{\tabcolsep}{0.5pt}
\renewcommand{\arraystretch}{0.5}
\begin{tabular}{c c c}
     \includegraphics[width=0.33\textwidth]{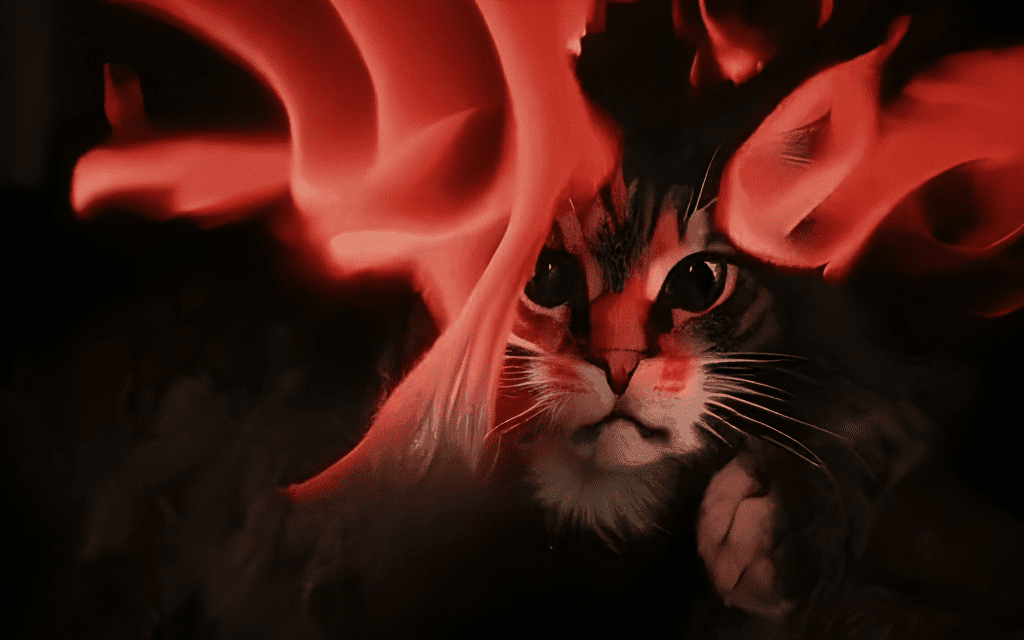} & 
      \includegraphics[width=0.33\textwidth]{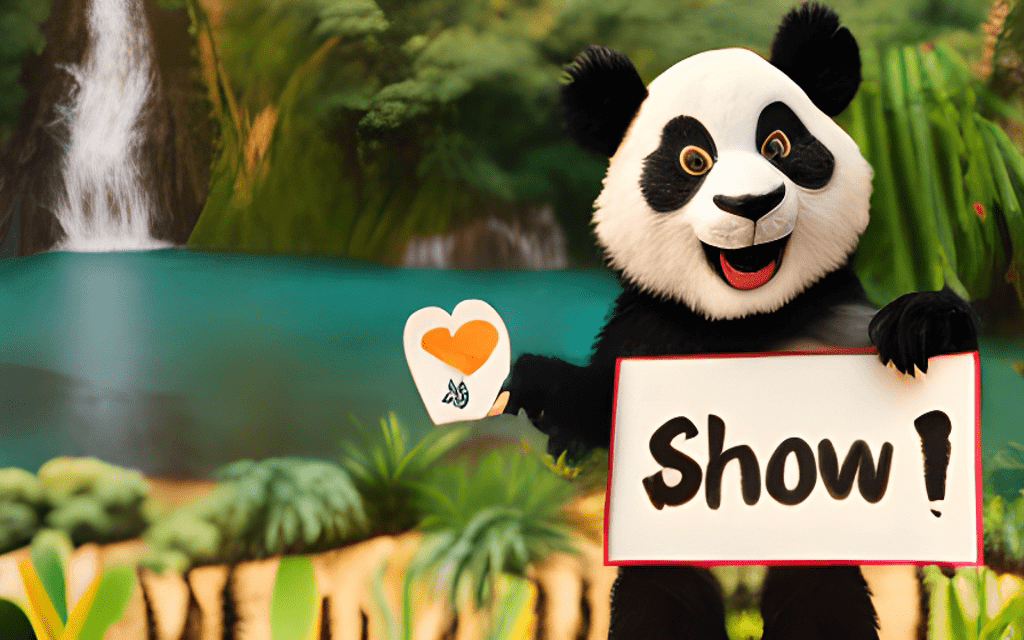} &
    \includegraphics[width=0.33\textwidth]{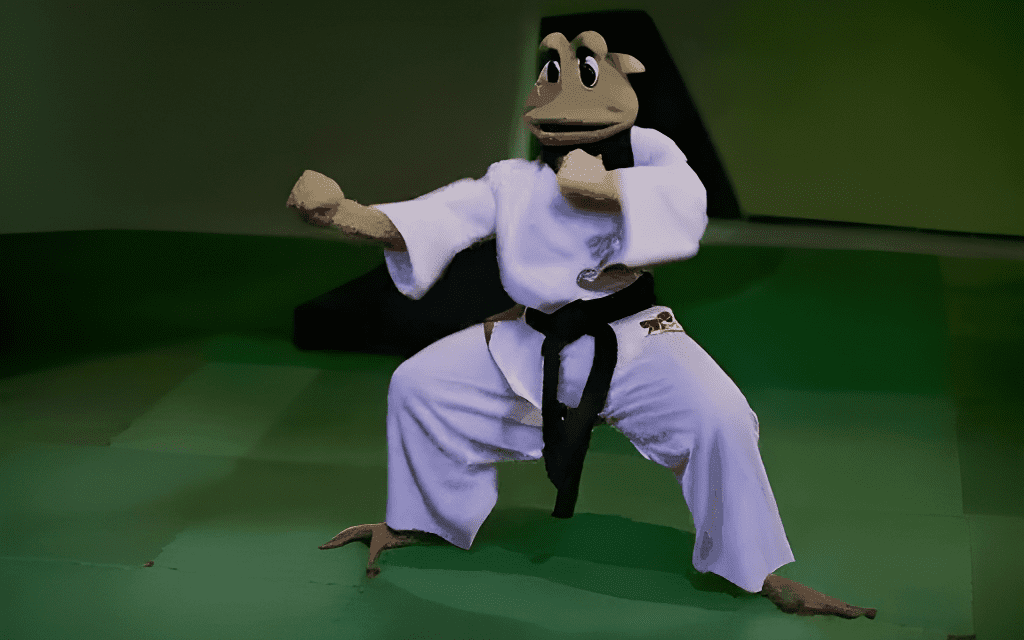}
\end{tabular}
\begin{tabularx}{\textwidth}{m{0.318\textwidth} c m{0.318\textwidth} c m{0.318\textwidth}} 
    \emph{\small Close up of mystic cat, like a buring phoenix, red and black colors.} & \hspace{0.6em} & 
    \emph{\small A panda besides the waterfall is holding a sign that says ``Show 1".} & \hspace{0.6em} &
    \emph{\small Toad practicing karate.} 
\end{tabularx}
\captionof{figure}{Given text descriptions, our approach generates highly faithful and photorealistic videos. \emph{\textcolor{magenta}{Click} the image to play the video clips. Best viewed with Adobe Acrobat Reader.}}
\vspace{8mm}
\label{fig:teaser}}]

\renewcommand{\thefootnote}{}
\footnote{ All authors are affiliated with Show Lab, National University of Singapore. David Junhao Zhang, Jay Zhangjie Wu and Jia-Wei Liu contribute equally. Mike Zheng Shou is the corresponding author.}

\renewcommand{\thefootnote}{\arabic{footnote}}
\begin{abstract}

Significant advancements have been achieved in the realm of large-scale pre-trained text-to-video Diffusion Models (VDMs). However, previous methods either rely solely on pixel-based VDMs, which come with high computational costs, or on latent-based VDMs, which often struggle with precise text-video alignment. In this paper, we are the first to propose a hybrid model, dubbed as Show-1, which marries pixel-based and latent-based VDMs for text-to-video generation. Our model first uses pixel-based VDMs to produce a low-resolution video of strong text-video correlation. After that, we propose a novel expert translation method that employs the latent-based VDMs to further upsample the low-resolution video to high resolution, which can also remove potential artifacts and corruptions from low-resolution videos. Compared to latent VDMs, Show-1 can produce high-quality videos of precise text-video alignment; Compared to pixel VDMs, Show-1 is much more efficient  (GPU memory usage during inference is 15G vs 72G). Furthermore, our Show-1 model can be readily adapted for motion customization and video stylization applications through simple temporal attention layer finetuning.  Our model achieves state-of-the-art performance on standard video generation benchmarks.  Code of Show-1 is publicly available and more videos can be found \href{https://junhaozhang98.github.io/show-1-ijcv/}{here}. 
\end{abstract}

\section{Introduction}

\begin{figure}[t]
    \centering
    \includegraphics[width=0.9\linewidth]{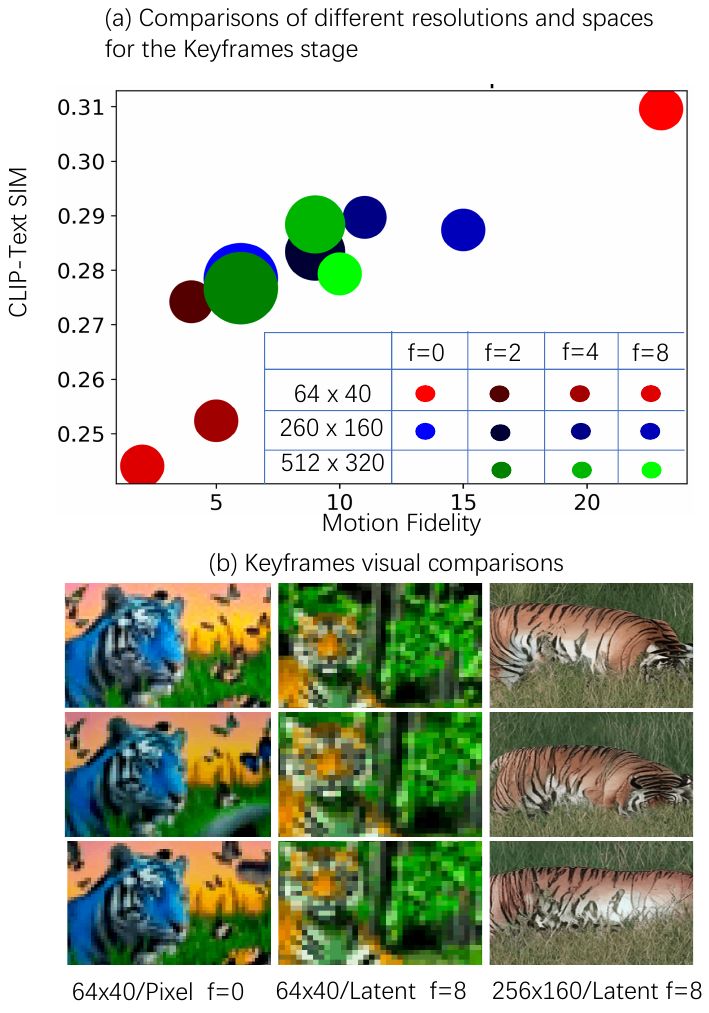}
    \caption{ The comparison (a) evaluates the CLIP-Text Similarity Score, highlighting how well the text aligns with video content and the fidelity of motion across various pixel and latent model pairings at different resolutions and compression ratios during the \textbf{keyframe stage}. These keyframe models all utilize identical latent VDM for the final super-resolution phases. The point's radius signifies the peak memory usage during the whole inference process. For consistency, all models in this study employ the same T5 text encoder and start with pre-trained weights from LAION, followed by additional training on WebVid using uniform steps to maintain fairness. $f=0$ indicates the model operating in pixel space, while $f=2,4,8$ correspond to different latent compression ratios.  The findings reveal that employing a pixel VDM to create low-resolution videos (64x40) at the keyframe stage yields superior outcomes compared to latent VDM across various resolutions and compression ratios. Part (b) presents the visual outcomes of the keyframes.}
    \label{fig:intro}
\end{figure}

Remarkable progress has been made in developing large-scale pre-trained Text-to-Video Diffusion Models (VDMs), including closed-source ones (\eg, Make-A-Video~\citep{singer2022make}, Imagen Video~\citep{ho2022imagen}, Video LDM~\citep{blattmann2023align}, Gen-2~\citep{esser2023structure}) and open-sourced ones (\eg, VideoCrafter~\citep{he2022latent}, ModelScopeT2V~\citep{wang2023modelscope}. 
These VDMs can be classified into two types: (1) Pixel-based VDMs that directly denoise pixel values, including Make-A-Video~\citep{singer2022make}, Imagen Video~\citep{ho2022imagen}, PYoCo~\citep{ge2023preserve}, and (2) Latent-based VDMs that manipulate the compacted latent space within a variational autoencoder (VAE), like Video LDM~\citep{blattmann2023align} and MagicVideo~\citep{zhou2022magicvideo}. 

However, both of them have pros and cons. As indicated by~\citep{singer2022make,ho2022imagen}, \textbf{pixel-based VDMs} can generate motion accurately aligned with the textual prompt because they start generating video from a very low resolution e.g., $64 \times 40$ (also demonstrated by Fig.~\ref{fig:intro}). But they typically demand expensive computational costs in terms of time and GPU memory, especially when upscaling the video to the  high-resolution.
\textbf{Latent-based VDMs} are more resource-efficient because they work in a reduced-dimension latent space.
But it is challenging for such small latent space (\eg, $8 \times 5$ for $64 \times 40$ videos) to cover rich yet necessary visual semantic details as described by the textual prompt.
Therefore, as shown in Fig.~\ref{fig:intro}, the generated videos often are not well-aligned with the textual prompts. On the other hand, when directly generating relatively high resolution videos (e.g., $256 \times 160$) using latent methods, the alignment between text and video could also be relatively weaker. This occurs because with higher resolution, the latent model tends to concentrate more on spatial appearance, potentially overlooking the text-video alignment, as validated by Fig.~\ref{fig:intro} and Tab.~\ref{ablation1}.

Prior models have often exclusively used either pixel or latent approaches across all above modules, facing the cons brought by either pixel or latent VDMs. Specifically, pure pixel-based VDMs e.g., Make-A-Video~\citep{singer2022make} are computationally demanding, while latent-based models may compromise text-video alignment and motion fidelity. To solve these problems, integrating the strengths of pixel-based and latent-based Video Diffusion Models (VDMs), while addressing their weaknesses, shows immense potential.  Achieving this integration could yield a text-to-video model that not only excels in video-text alignment but also with low computation cost.  

\begin{figure}[t]
    \centering
    \includegraphics[width=0.9\linewidth]{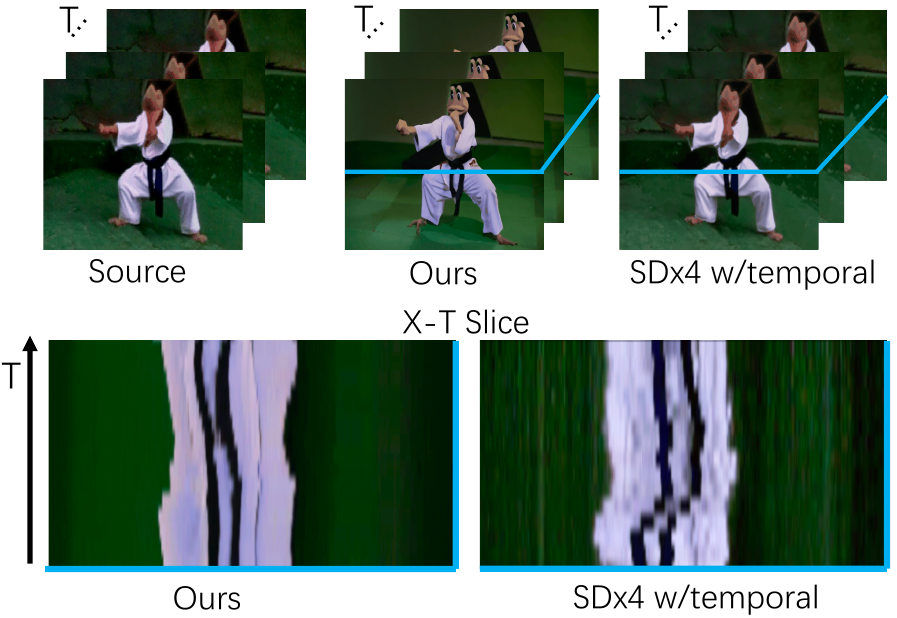}
    \caption{\textbf{Final Super-Resolution Comparisons.} We contrast our expert translation against typical SDx4 upsampling that includes temporal layers and visualize the X-T slice of the final outcomes. The findings suggest that our approach is capable of managing the possible corruptions found in low-resolution videos, resulting in improved temporal consistency and quality (notably smoother and with reduced noise in the X-T slice) compared to SDx4 with temporal layers.
}
    \label{fig:intro2}
\end{figure}

Toward this objective, we begin a step-by-step exploration of how to merge pixel and latent VDMs effectively. Referencing Fig.~\ref{fig:intro}, we observe that initiating video generation with low-resolution  keyframes using pixel-based VDM leads to improved text-video alignment. Accordingly, we employ a coarse-to-fine generation strategy that starts by creating low-resolution and low-frame-rate keyframes using pixel-based  VDM. Then we apply a temporal interpolation module and a super-resolution module to enhance the video in both time and space. In the current step, we leverage the advantages of pixel VDMs, resulting in an improved text-aligned low-resolution video. 

However, as previously mentioned, continuing to use pixel VDMs as the final super-resolution module for ultimate high-resolution output will result in significant computational costs. Thus, we opt for a latent-based VDM for an efficient final super-resolution module. Typical latent-based VDMs, such as SDx4~\citep{rombach2022high}, usually combine low-resolution video and noise as input for a UNet. Nonetheless, as shown in Fig.~\ref{fig:intro2}, there might be some artifacts or corruptions originating from the low-resolution videos. Simply applying typical latent-based VDMs like SDx4 with a temporal extension will not address these issues, leading to subpar final results and poor temporal consistency, as evidenced by the discontinuous and noisy X-T slice in Fig.~\ref{fig:intro2}. To overcome this problem, we introduce an expert translation method for latent-based VDMs, which directly uses the encoded noisy low-resolution video as the input for UNet with expert finetuning. We discover that latent-based VDMs with expert translation can effectively convert low-resolution video to high-resolution while preserving the original appearance and accurate text-video alignment. Crucially, it also eliminates the artifacts and corruptions from low resolution videos.

Ultimately, we successfully integrate the benefits of both pixel and latent-based VDMs within a cohesive framework, named Show-1, which achieves state-of-the-art performance  on popular video generation benchmarks including UCF101~\citep{soomro2012ucf101}, MSR-VTT~\citep{xu2016msr} and VBench~\citep{huang2023vbench}. Additionally, by exclusively fine-tuning the temporal attention layers of the keyframes UNet on a single video, Show-1 is capable of distilling the video's motion into these layers. This process allows for motion customization and stylization of the video, as the fixed spatial layers offer a range of appearances based on the text.

\begin{figure*}[t]
    \centering
    \includegraphics[width=\linewidth]{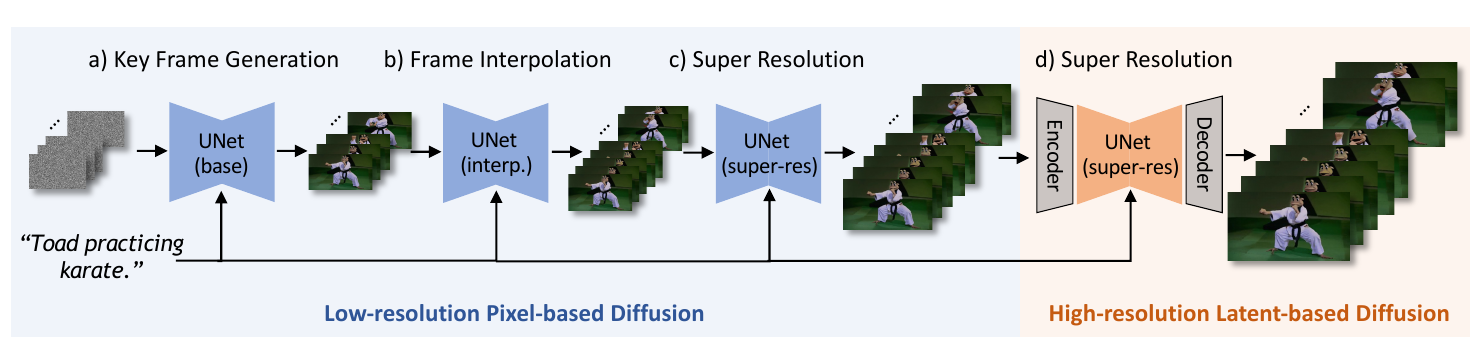}
    \caption{\textbf{Overview of Show-1}. Pixel-based VDMs produce videos of lower resolution with better text-video alignment, while latent-based VDMs upscale these low-resolution videos from pixel-based VDMs to then create high-resolution videos with low computation cost.}
    \label{fig:architecture}
\end{figure*}



The key contributions of our paper is summarized as follows:

\begin{itemize}
    \setlength\itemsep{1mm}
    \item Upon examining pixel and latent VDMs, we discover that: 1) pixel VDMs excel in generating low-resolution videos with more natural motion and superior text-video synchronization compared to latent VDMs; 2) when using the low-resolution video as an initial guide, latent VDMs can effectively function as super-resolution tools by \textbf{simple expert translation}, refining spatial clarity and creating high-quality videos with greater efficiency than pixel VDMs. Meanwhile, with expert translation, the artifacts and corruptions of low resolution videos can be reduced.
   
    \item We are the first to integrate the strengths of both pixel and latent VDMs, resulting into a novel video generation model that can produce high-resolution videos of precise text-video alignment at low computational cost (15G GPU memory during inference).

    \item By fine-tuning the temporal attention layer, our Show-1 model can be additionally adapted for motion customization and video stylization applications.


\end{itemize}

\section{Related Work}

\paragraph{Text-to-image generation.} \citep{reed2016generative} stands as one of the initial methods that adapts the unconditional Generative Adversarial Network (GAN) introduced by \citep{goodfellow2014generative} for text-to-image (T2I) generation. Later versions of GANs delve into progressive generation, as seen in \citep{zhang2017stackgan} and \citep{hong2018inferring}. Meanwhile, works like \citep{xu2018attngan} and \citep{zhang2021cross} seek to improve text-image alignment. Recently, diffusion models have contributed prominently to advancements in text-driven photorealistic and compositional image synthesis~\citep{ramesh2022hierarchical,saharia2022photorealistic}.   For attaining high-resolution imagery, two prevalent strategies emerge. One integrates cascaded super-resolution mechanisms within the RGB domain~\citep{nichol2021glide,ho2022cascaded,saharia2022photorealistic,ramesh2022hierarchical}. In contrast, the other harnesses decoders to delve into latent spaces~\citep{rombach2022high,gu2022vector}.  Owing to the emergence of robust text-to-image diffusion models, we are able to utilize them as solid initialization
 of text to video models.
 
\paragraph{Text-to-video generation.} 
Past research has utilized a range of generative models, including GANs~\citep{vondrick2016generating,saito2017temporal,Tulyakov_2018_CVPR,tian2021a,Shen_2023_CVPR}, autoregressive models~\citep{srivastava2015unsupervised,yan2021videogpt,le2021ccvs,ge2022long,hong2022cogvideo,kondratyuk2023videopoet}, and implicit neural representations~\citep{skorokhodov2021stylegan,yu2021generating}. Inspired by the notable success of the diffusion model in image synthesis, several recent studies have ventured into applying diffusion models for both conditional and unconditional video synthesis~\citep{voleti2022masked,harvey2022flexible,zhou2022magicvideo,wu2022tune,blattmann2023videoldm,khachatryan2023text2video,hoppe2022diffusion,voleti2022masked,yang2022diffusion,nikankin2022sinfusion,luo2023videofusion,an2023latent,wang2023videofactory}. Several studies have investigated the hierarchical structure, encompassing separate keyframes, interpolation, and super-resolution modules for high-fidelity video generation. Magicvideo~\citep{zhou2022magicvideo}, VideoFactory~\citep{wang2023videofactory}, NUWA-XL~\citep{yin2023nuwa}, LaVie~\citep{wang2023lavie}, VideoCrafter~\citep{chen2023videocrafter1} and Video LDM~\citep{blattmann2023align} ground their models on latent-based VDMs. On the other hand, PYoCo~\citep{ge2023preserve}, Make-A-Video~\citep{singer2022make}, Lumiere~\cite{bar2024lumiere} and Imagen Video~\citep{ho2022imagen} anchor their models on pixel-based VDMs. These methods primarily rely on either pixel-based VDM or latent-based VDM. Using only pixel-based VDM results in improved text-video alignment and motion fidelity, but at the expense of significant computational resources. On the other hand, relying solely on latent-based VDM is more efficient, yet it presents challenges in achieving high-quality text-video alignment and motion fidelity. Unlike these methods, our approach investigates how to effectively combine pixel-based and latent-based VDMs, leveraging the strengths and avoiding the weaknesses of both pixel-based and latent-based VDMs.

\vspace{-3mm}

\section{\textit{Show-1}}
\subsection{Preliminaries}

\paragraph{Denoising Diffusion Probabilistic Models (DDPMs)~\citep{ho2020denoising}}  are generative models that utilize a reverse Markov chain to synthesize data, beginning from a noise distribution and progressively denoising it. This process is driven by optimizing model parameters to align the reverse sequence with the forward noisy sequence. The training objective focuses on minimizing the difference between the actual noise and the noise estimated by the model, formalized as follows:
\begin{equation}
   \quad\quad\quad\quad\quad\quad \mathbb{E}_{x, \epsilon \sim \mathcal{N}(0, 1), t} \left[ \lVert \epsilon - \epsilon_\theta(x_t, t) \rVert_2^2 \right].
\end{equation}

This expression represents the expected value of the squared $L2$ norm between the noise \(\epsilon\) and the noise predicted by the model \(\epsilon_{\theta}\), where \(\epsilon\) is drawn from a standard Gaussian distribution and \(x_t\) is the noisy data at timestep \(t\). The model's parameters \(\theta\) are trained to minimize this expectation, which corresponds to denoising the data point \(x_t\).

\paragraph{UNet architecture for text-to-image model.}
The UNet model is first introduced by ~\citep{2015u} for biomedical image segmentation. Popular UNet for text-to-image diffusion model usually contains multiple down, middle, and up blocks. Each block consists of a ResNet2D layer, a self-attention layer, and a cross-attention layer. The cross-attention layers play a crucial role in fusing images and texts, allowing text-to-image models to generate images that are consistent with textual descriptions. Text condition $c$ is inserted into cross-attention layer as keys and values. For a text-guided diffusion model, with the text embedding $c$, the objective is given by:
\begin{equation}
 \quad\quad\quad\quad\quad \mathbb{E}_{x, \epsilon \sim \mathcal{N}(0, 1), t, c} \left[ \| \epsilon - \epsilon_\theta(x_t, t, c) \|^2_2 \right].
\label{objective}
\end{equation}

\subsection{Turn Image UNet to Video}  
\begin{figure}[t]
    \centering
    \includegraphics[width=0.9\linewidth]{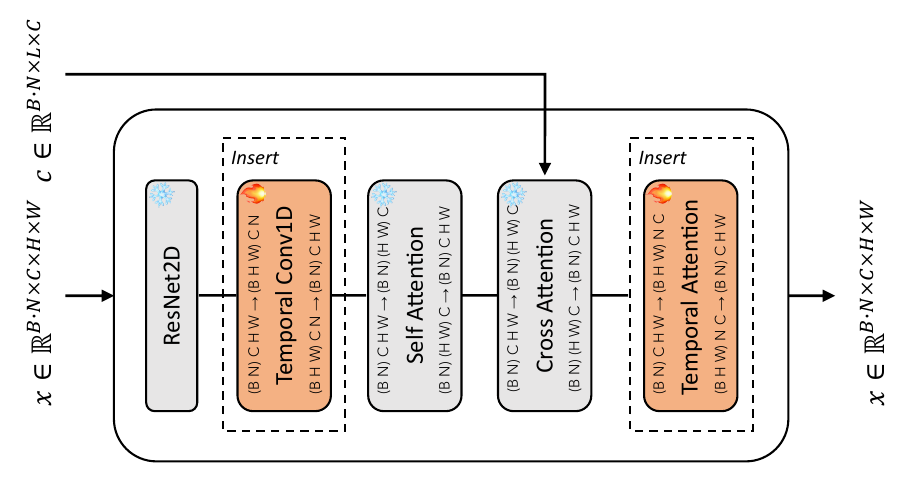}
    \vspace{-3mm}
    \caption{\textbf{UNet block of Show-1.} We modify the 2D UNet by inserting temporal convolution and attention layers inside each block. During training, we update the additional temporal layers while keeping spatial layers fixed.}
    \label{fig:block}
\end{figure}

We use the spatial weights from a robust text-to-image model. To endow the model with temporal understanding and produce coherent frames, as shown in Fig.~\ref{fig:block}, we integrate temporal layers within each UNet block. Specifically, after every Resnet2D block, we introduce a temporal convolution layer consisting of four 1D convolutions across the temporal dimension. Additionally, following each spatial self- and cross-attention layer, we implement a temporal attention layer to facilitate dynamic temporal data assimilation. 
Formally, given a frame-wise input video $x \in \mathcal{R}^{N \times C \times H \times W}$,
where $N$ is number of frames, $C$ is the number of channels, $H$ and $W$ are the spatial latent dimensions,
the spatial self-attention layer operates the input video as a sequence of independent spatial images by transposing the temporal axis into the batch dimension, as illustrated below using \texttt{einops}~\citep{rogozhnikov2022einops} (Here, we include the batch size $B$ to better illustrate the transpose operation. After this, we omit $B$ for simplicity in notation.):
\[
x_{\text{SA}} \leftarrow \texttt{rearrange}(x,\texttt{(B N C H W} \rightarrow \texttt{(B N) (H W) C}). \\
\]
For temporal self-attention layer, the video is reshaped back to temporal dimensions:
\[
x_{\text{TA}} \leftarrow \texttt{rearrange}(x,\texttt{(B N C H W} \rightarrow \texttt{(B H W) N C}). \\
\]
The attention mechanism~\citep{vaswani2017attention} implements $\mathrm{Attention}(Q,K,V)=\mathrm{Softmax}(\frac{Q K^T}{\sqrt{d}}) \cdot V$, with 
{$$Q=W_Q x, K=W_K x, V=W_V x,$$}%
where $W_Q$, $W_K$, and $W_V$ are learnable matrices that project the inputs to query, key and value, respectively, and $d$ is the output dimension of key and query features. The $x$ is transposed to $x_{\text{SA}}$ and $x_{\text{TA}}$ for spatial and temporal self-attention respectively. Differently, the cross-attention layer receives key and value matrices from the text prompt:
{$$Q=W_Q x_{\text{SA}}, K=W_K c, V=W_V c,$$}%
where $c \in \mathcal{R}^{N \times L \times C}$ is the encoded text embedding and $L$ denotes the sequence length of text embedding.

\subsection{Pixel-based Keyframe Generation Model}\label{sec:keyframe}
Given a text input, we initially produce a sequence of keyframes using a pixel-based Video UNet at a very low spatial and temporal resolution (Fig.~\ref{fig:architecture}, Stage a. This approach results in improved text-to-video alignment. The reason for this enhancement is that we do not require the keyframe modules to prioritize appearance clarity or temporal consistency given that the resolution of video is very low. As a result, the keyframe modules pay more attention to the text guidance. The training objective for the keyframe modules is following Eq.~\ref{objective}.

\paragraph{Why we choose pixel diffusion over latent diffusion here?}
\textbf{1)}Latent diffusion employs an encoder to transform the original input $x$ into a latent space. This results in a reduced spatial dimension, for example, $H/8, W/8$, while concentrating the semantics and appearance into this latent domain. For generating keyframes, our objective is to have a smaller spatial dimension, like $64 \times 40$. If we opt for latent diffusion, this spatial dimension would shrink further to around $ 8 \times 5$, which is not be sufficient to retain ample spatial semantics and appearance within the compacted latent space, resulting in poor text-video alignment as shown in Tab.~\ref{ablation1}. On the other hand, pixel diffusion operates directly in the pixel domain, keeping the original spatial dimension intact. This ensures that  necessary semantics and appearance information are preserved. For the following low resolution stages, we all utilize pixel-based VDMs for the same reason. \textbf{2)} An alternative is to lower the compression ratio of latent VDMs. Yet, as highlighted in ~\citep{rombach2022high}, latent diffusion's main goal is to cut down on computational and memory demands significantly. For instance, stable diffusion compresses a $512 \times 512$ image to a $64 \times 64$ latent size, achieving 8-fold reduction. However, with a minimal compression ratio, like 2-fold, the efficiency and training costs become comparable to pixel diffusion, as stated in ~\citep{rombach2022high}. Thus, with a low compression ratio, latent diffusion may be unnecessary, especially since it requires training an extra autoencoder, whereas pixel diffusion does not. \textbf{3)} Another approach involves using latent-based VDM to generate high-resolution keyframes. However, as indicated in Tab.~\ref{ablation1}, directly generating high-resolution keyframes leads to poorer text-video alignment and motion quality compared to generating low-resolution keyframes with pixel-based VDM. Furthermore, as the resolution increases ($512 \times 320$ vs $256 \times 160$ in Tab.~\ref{ablation1}), both text-video alignment and motion fidelity deteriorate. These findings suggest that at higher resolutions, the latent model may focus more on spatial appearance, potentially neglecting text-video alignment and motion fidelity.

\subsection{Temporal Interpolation Model}\label{sec:interp}

We enhance the temporal resolution of videos with a pixel-based temporal interpolation diffusion model (Fig.~\ref{fig:architecture}, Stage b), which iteratively predicts the intermediate frames between the keyframes produced by the previous keyframe model (Sec.~\ref{sec:keyframe}). 
We employ the masking technique, as highlighted in ~\citep{blattmann2023align}, where the target intermediate frames to be interpolated are masked during training process. 
We inherit the UNet architecture from keyframe model and modify the input channels of the first convolution layer to accommodate the masked key frames as condition via channel-wise concatenation. 
Specifically, we start from the noisy video frames segment $\{x_{t}^i, x_{t}^{j}, x_{t}^{j+1}, x_{t}^{j+2}, x_{t}^{i+1}\} \in \mathcal{R}^{5 \times C \times H \times W}$ at timestep $t$, where $x_{t}^{\{i, i+1\}}$ are two consecutive key frames and $x_{t}^{\{j, j+1, j+2\}}$ are three intermediate frames to be interpolated. As depicted in Fig.~\ref{fig:input} (Interpolation), we concatenate them with the original key frames $x_0 \in \mathcal{R}^{5 \times C \times H \times W}$ and addition binary masks $m \in \mathcal{R}^{5 \times 1 \times H \times W}$ along the channel dimension as conditioning signals, resulting in an input shape of ${5 \times (C+C+1) \times H \times W}$. We set $x_0^{\{j, j+1, j+2\}}$ and $m^{\{j, j+1, j+2\}}$ to 0, indicating the frames to be interpolated. Note that $x_0$ and $m$ serve as the conditions. The UNet takes the concatenation with the shape of ${5 \times (C+C+1) \times H \times W}$ as its input. Then the UNet outputs noise with a shape of ${5 \times C \times H \times W}$ as the prediction of the noise at timestep $t$ for $\{x_{t}^i, x_{t}^{j}, x_{t}^{j+1}, x_{t}^{j+2}, x_{t}^{i+1}\} \in \mathcal{R}^{5 \times C \times H \times W}$.
We apply noise conditioning augmentation to conditional key frames \(x_{0}^{i}\) and \(x_{0}^{i+1}\) by adding a small amount of random noise. Such augmentation is pivotal in cascaded diffusion models for conditional generation, as observed by ~\citep{ho2022imagen}, and also in text-to-image models as noted by ~\citep{he2022latent}. It aids in the simultaneous training of diverse models in a cascade manner and minimizes the vulnerability to domain disparities between the output from previous phase and the training inputs of the following phase. Let the interpolated video frames be represented by $x^{\prime} \in \mathcal{R}^{4N \times C \times H \times W}$($x_{t}^{\prime}$ can be regarded as the combination of multiple overlap segments $\{x_{t}^i, x_{t}^{j}, x_{t}^{j+1}, x_{t}^{j+2}, x_{t}^{i+1}\}$. Here we use $4N$ instead of $4N-3$ for simpler notation). Based on Eq.~\ref{objective}, we can formulate the updated objective as:
{\begin{equation}
\quad \quad  \mathbb{E}_{x^{\prime}, x_0, m ,\epsilon \sim \mathcal{N}(0, 1), t, c} \left[ \| \epsilon - \epsilon_\theta([x^{\prime}_t,x_0,m], t, c) \|^2_2 \right].
\end{equation}
\label{inter}}%
Notably, we reuse the pretrained weights of keyframe model, exluding the last four channels of the first convolution layer, to finetune the interpolation model for fast convergence.

\vspace{-5mm}
\subsection{Super-resolution at Low Spatial Resolution}

Upscaling a low-resolution video by $8\times$ presents a significant challenge for a single super-resolution module, given that a video with low resolution, such as one with dimensions of 
$64\times 40$, lacks sufficient visual detail. To address this, we divide the super-resolution process into two distinct modules. The initial module is tasked with enhancing the spatial quality of the low-resolution video, while the subsequent module is dedicated to generating the final high-resolution output.

\begin{figure}[t]
    \centering
    \includegraphics[width=0.9\linewidth]{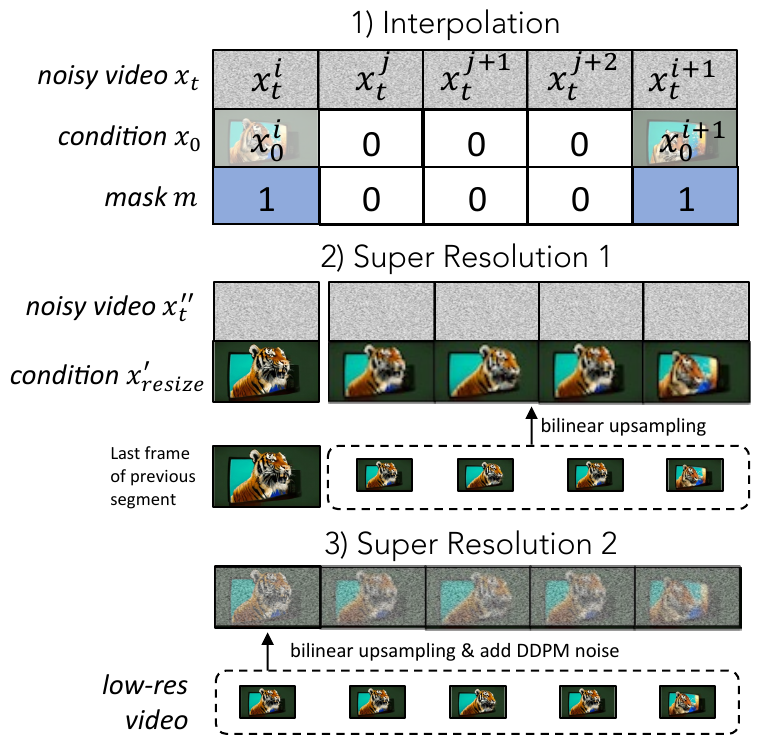}
    \vspace{-3mm}
    \caption{\textbf{Illustration of the input for interpolation and super-resolution modules.} \textit{Interpolation:} We concatenate noise with low-FPS frames and a mask that indicates the conditional frames. \textit{Super Resolution 1:} We resize the low-resolution frames to high-resolution using bilinear upsampling and concatenate them with input noise. We also use the last frame of the previous segment as a condition to enable autoregressive upsampling. \textit{Super Resolution 2:} We resize the input video to high-resolution, and follow SDEdit~\citep{meng2021sdedit} to add DDPM noise and gradually remove it.}
    \label{fig:input}
\end{figure}

In the first low resolution video upsampling module (Fig.~\ref{fig:architecture}, Stage c), we introduce a pixel super-resolution approach utilizing the video UNet. The super-resolution model takes as input a low-resolution video $x^{\prime} \in \mathcal{R}^{4N \times C \times H \times W}$ produced by previous stages and outputs a high-resolution video $x^{\prime\prime} \in \mathcal{R}^{4N \times C \times 4H \times 4W}$ with a $4\times$ increase in spatial dimension. 
Similar to the channel-wise conditioning in interpolation model (Sec.~\ref{sec:interp}), we concatenate the noisy video frames $x_t^{\prime\prime} \in \mathcal{R}^{4N \times C \times 4H \times 4W}$ at the timestep $t$ with the resized low-resolution video clip $x_{resized}^{\prime} \in \mathcal{R}^{4N \times C \times 4H \times 4W}$, which is bilinearly upsampled to fit the spatial size of high-resolution video ( Fig.~\ref{fig:input}, Super Resolution 1).
The UNet takes the concatenation $[x_t^{\prime\prime},x_{resized}^{\prime}]$ with the shape of $4N \times (C+C) \times 4H \times 4W$ as its input. Then the UNet outputs noise with a shape of $4N \times C \times 4H \times 4W$ as the prediction of the noise at timestep $t$ for $x_{t}^{\prime\prime}\in \mathcal{R}^{4N \times C \times 4H \times 4W}$.
In line with the approach Imagen Video~\citep{ho2022imagen}, we employ gaussian noise augmentation to the upscaled low resolution video condition during its training process, introducing a random signal-to-noise ratio. This augmentation can minimize the domain gap between the output from previous interpolation stage and the training inputs of the following stage. During the sampling process, we opt for a consistent signal-to-noise ratio, like 1 or 2. 
Meanwhile, given that the spatial resolution remains at an upscaled version throughout the diffusion process, it's challenging to upscale all the interpolated frames in one forward process using a standard GPU with 24GB memory. Consequently, we must divide the frames into four smaller segments and upscale each one individually.  
However, the continuity between various segments is compromised. To rectify this, as depicted in the Fig.~\ref{fig:input}, we take the upscaled last frame of one segment to complete the three supplementary channels of the initial frame in the following segment.

\subsection{Super-resolution at High Spatial Resolution}
\label{section3.6}

Sometimes, previous stages may generate videos with artifacts or temporal corruptions as shown in Fig.~\ref{fig:intro2}. Therefore, the final super-resolution module \textcolor{blue}{(Fig.~\ref{fig:architecture}, Stage d)} is tasked with managing these artifacts to produce videos of high quality as the end result.

However, injecting the low-resolution input via channel-wise concatenation, as per the first super-resolution module and SDx4\footnote{https://huggingface.co/stabilityai/stable-diffusion-x4-upscaler}, is inadequate for addressing these artifacts. This approach results in the poor temporal consistency for high spatial resolution, as demonstrated in the X-T slice of Fig.~\ref{fig:intro2}. To overcome this issue, we introduce an expert translation for a latent-based Video Diffusion Model (VDM), which proves to be effective in higher resolution stages. This involves two key modifications from the SDx4 approach. \textbf{Firstly,} we implement a noising-denoising process, as outlined by \emph{SDEdit}~\citep{meng2021sdedit}, on the encoded low-resolution videos from earlier stages. These processed videos serve as the input for the UNet, and we do this without appending any extra channels. Specifically, SDEdit utilizes the pretrained diffusion with timesteps ranging from 0 to 1000 but begins inference from a noisy input at an intermediate timestep.  Inspired by this, we take low-resolution videos from previous stages, linearly interpolate them to a higher resolution, and add noise at an intermediate timestep. Then we apply the diffusion process from this intermediate timestep to 0 with the same prompt, resulting in more detailed outputs than the original linear interpolation.

  \textbf{Secondly,} as pointed out by \citep{balaji2022ediffi}, various diffusion steps assume distinct roles during the generation process. For instance, the initial diffusion steps, such as from 1000 to 900, primarily concentrate on recovering the overall spatial structure, while subsequent steps delve into finer details. Given our success in securing well-structured low-resolution videos, we suggest adapting the latent VDM to specialize in high-resolution detail refinement. More precisely, we train a UNet for only the 0 to 900 timesteps (with 1000 being the maximum) instead of the typical full range of 0 to 1000, directing the model to be an expert emphasizing high-resolution nuances. This strategic adjustment significantly enhances the end video quality, namely expert finetuning. With our first SDEdit modification, we can perform the denoising process from an intermediate timestep, such as 900, for the noisy linearly interpolated  video. Therefore, the loss of knowledge from timesteps 1000 to 900 due to expert fine-tuning is not an issue. Through our empirical observations, we discern that a latent-based VDM with our expert translation can be effectively utilized for enhanced super-resolution with high fidelity and great temporal consistency. This results in the final video, denoted as 
  \(x^{\prime\prime\prime} \in \mathcal{R}^{4N \times C \times 8H \times 8W}\).  
  
\paragraph{Why choose latent-based VDM over pixel-based VDM here?}  Pixel-based VDMs work directly within the pixel domain, preserving the original spatial dimensions. Handling high-resolution videos this way can be computationally expensive. As shown in Tab.~\ref{ablation1}, using pixel-based VDM for final super-resolution requires huge GPU memory e.g., 72GB.  In contrast, latent-based VDMs compress videos into a latent space (for example, downscaled by a factor of 8), which results in a reduced computational burden. Moreover, although latent-based VDM may result in less precise text-video alignment, it can
be re-purposed to translate low-resolution video to high-resolution video, while maintaining the
original appearance and the accurate text-video alignment of low-resolution video generated by the pixel base-VDM. Thus, we opt for the latent-based VDMs here. 

Another choice is to reduce the parameters of  pixel model. For example, Make-A-Video~\citep{singer2022make} reduces its final superresolution model to 0.7B parameters. However, it  still requires substantial computational costs because its UNet  directly operates on the high output resolution. We replicate Make-A-Video with its original parameters and architecture (Tab.~
\ref{table:speed}) and find that  even with 0.7B parameters, it's still computationally demanding with 52GB memory, while our latent upsampling only needs 15G. Moreover, upsampling synthetic videos also poses challenges, particularly due to the domain gap between training on real data and testing on synthetic outputs. Achieving temporal consistency and high visual quality while minimizing artifacts  requires high  model complexity, which is impractical with further reducing parameters.

\subsection{Motion Customization and Video Stylization.} Drawing from recent advancements in Motion Customization~\citep{zhao2023motiondirector,jeong2023vmc}, we have further developed our model to accommodate these applications. In contrast to the Motion Director~\citep{zhao2023motiondirector} approach, which requires separate training for spatial and temporal layers tailored to a specific video, our method stands out by focusing fine-tuning efforts solely on the temporal attention layers of the keyframes' UNet. This targeted approach to fine-tuning is designed to be computationally and memory-efficient. Through this process, we are able to encapsulate the motion dynamics of the given video within the temporal attention layers.  It's important to highlight that the later modules for frame interpolation and spatial super-resolution are left unchanged. This approach allows for flexible video editing/ stylization and  tailored adjustments while maintaining the model's fundamental capability for broad synthesis.
\label{section3.7}

\section{Experiments}

\subsection{Implementation Details}

For the generation of pixel-based keyframes, we produce videos of dimensions $8 \times 64 \times 40 \times 3 (N \times H \times W \times 3)$. In our interpolation model, we initialize the weights using the keyframes generation model and produce videos with dimensions of $29 \times 64 \times 40 \times 3$. For  the first superresolution module, we upsample the video yielding the size $29 \times 256 \times 160$. In the subsequent super-resolution module, we modify the  latent-based VDM  and use our proposed expert translation to generate videos of $29 \times 576 \times 320$.

\begin{table}[t]
\centering
\caption{\textbf{Zero-shot text-to-video generation on UCF-101. Ours  achieves competitive results  in   inception score and FVD metrics.}}
\label{ucf1012}
\resizebox{0.45\textwidth}{!}{
\begin{tabular}{lcc}
\hline
Method              & IS  $(\uparrow)$   & FVD  $(\downarrow)$   \\
\midrule
CogVideo~\citep{hong2022cogvideo} (English) & 25.27 & 701.59 \\
Make-A-Video~\citep{singer2022make}         & 33.00 & \textbf{367.23} \\
MagicVideo~\citep{zhou2022magicvideo}   & - & 655.00 \\
Video LDM~\citep{blattmann2023align}   & 33.45 & 550.61 \\
VideoFactory~\citep{wang2023videofactory}  & - & 410.00 \\
\hline

Show-1 (ours) resized     & \underline{35.67} &  383.46 \\
Show-1 (ours)  finetune on square videos    & \textbf{36.02} &  \underline{369.33} \\
\hline
\end{tabular}}
\end{table}

In terms of training, we employ the public WebVid-10M dataset~\citep{bain2021frozen} as our video training data. Our infrastructure comprised 64 A100 GPUs, each with 40GB, which stands in contrast to the setups used by LaVie~\citep{wang2023lavie}, ModelScope~\citep{wang2023modelscope}, or~\citep{chen2023videocrafter1} VideoCrafter. These methods train on large scale internal datasets with more than 128 A100 GPUs, each with 80GB, which require \textbf{much more data and training resources than ours}.

Regarding the ablation studies depicted in Fig.~\ref{fig:intro}, Tab.~\ref{ablation1} and Tab.~\ref{tab:ablation2}, we ensure that the same T5 text encoder~\citep{2020t5}  is employed across both pixel-based and latent-based VDMs in the keyframes stage. Each model is initialized with the image model weights pre-trained on the LAION~\citep{schuhmann2022laion} dataset and has the same number of parameters, maintaining consistency for fair comparisons. Regarding comparisons with the state-of-the-art, our choice of initialization for the pre-trained Text-to-Image model is DeepFloyd\footnote{https://github.com/deep-floyd/IF}, which serves as the foundation for our model training.

\subsection{Quantitative Results}

\paragraph{UCF-101 Experiment.} For our preliminary evaluations, we employ IS and FVD metrics. UCF-101 stands out as a categorized video dataset curated for action recognition tasks. When extracting samples from the text-to-video model, following PYoCo \citep{ge2023preserve}, we formulate a series of prompts corresponding to each class name, serving as the conditional input. This step becomes essential for class names like \textit{jump rope}, which aren't intrinsically descriptive. Following ~\citep{singer2022make}, we generate totally 10000 video samples to determine the IS metric. For FVD evaluation, we adhere to methodologies presented in prior studies~\citep{le2021ccvs,tian2021a} and produce 2,048 videos.  To ensure a fair comparison with the previous methods \citep{ge2022long, singer2022make}, which produces square videos since it is directly trained on squared videos, we directly resized our generated videos to square videos. However, this resizing process introduces slight performance degradation to our model. We believe that a more rigorous approach would involve fine-tuning our entire pipeline on square videos to better align with the comparison criteria. Consequently, we present the results for both the resized version and the version fine-tuned on square videos in Table~\ref{ucf1012}

\begin{table}[!t]
\centering

\caption{
\textbf{Comparisons with SOTA  models on MSR-VTT dataset~\citep{xu2016msr}.}}

\label{mrtt}
\resizebox{0.48\textwidth}{!}{
\begin{tabular}{c|ccc}
\hline
\textbf{Models}              & FID-vid ($\downarrow$)   &  FVD ($\downarrow$) & CLIPSIM ($\uparrow$) \\ \hline
N\"UWA~\citep{wu2022nuwa}       & 47.68   & -  & 0.2439        \\
CogVideo (Chinese)~\citep{hong2022cogvideo}   & 24.78 & - & 0.2614      \\
CogVideo (English)~\citep{hong2022cogvideo}   & 23.59  & 1294 & 0.2631  \\
MagicVideo~\citep{zhou2022magicvideo}          & -    & 1290 & -        \\
Video LDM~\citep{blattmann2023align}      & -   &  - & 0.2929 \\ 
Make-A-Video~\citep{singer2022make}        & 13.17 &  -& \underline{0.3049}     \\ 
ModelScopeT2V~\citep{wang2023modelscope}  &   \textbf{11.09} & \underline{550} & 0.2930    \\ 
\hline

Show-1(ours) &  \underline{12.97} & \textbf{536} & \textbf{0.3104}   \\ \hline
\end{tabular}}
\end{table}

From the data presented in Tab.~\ref{ucf1012}, it's evident that Show-1's zero-shot capabilities outperform or are on par with other methods. This underscores Show-1's superior ability to generalize effectively, even in specialized domains. It's noteworthy that our keyframes, interpolation, and initial super-resolution models are solely trained on the publicly available WebVid-10M dataset, in contrast to the Make-A-Video models, which are trained on large scale internal text-video data.

\begin{table*}[ht]
\centering
\caption{\textbf{VBench Evaluation Results per Dimension.} This table compares the performance of five video generation models across each of the 16 VBench dimensions. A higher score indicates relatively better performance for a particular dimension. }
\label{tab:vbench}
\resizebox{0.95\linewidth}{!}{
\begin{tabular}{c|c|c|c|c|c|c|c|c}
\hline
Models & \Centerstack{Subject\\Consistency} & \Centerstack{Background\\Consistency} & 
\Centerstack{Temporal\\Flickering} & 
\Centerstack{Motion\\Smoothness} & 
\Centerstack{Dynamic\\Degree} & 
\Centerstack{Aesthetic\\Quality} &  
\Centerstack{Imaging\\Quality}&
\Centerstack{Object\\Class}\\ \hline
LaVie      & 91.41\% & 97.47\% & 98.30\% & 96.38\% & 49.72\% & 54.94\% & \textbf{61.90\%} & 91.82\% \\ 
ModelScope  & 89.87\% & 95.29\% & 98.28\% & 95.79\% & 66.39\% & 52.06\% & 58.57\% & 82.25\% \\ 
VideoCrafter & 86.24\% & 92.88\% & 97.60\% & 91.79\% & \textbf{89.72\%} & 44.41\% & 57.22\% & 87.34\% \\ 
CogVideo & 92.19\% & 95.42\% & 97.64\% & 96.47\% & 42.22\% & 38.18\% & 41.03\% & 73.40\% \\ \hline
Show-1 & \textbf{95.53}\% & \textbf{98.02}\% & \textbf{99.12}\% & \textbf{98.24}\% & 44.44\% & \textbf{57.35}\% & 59.75\% & \textbf{93.07}\%  \\
\hline
\hline
Models   & \Centerstack{Multiple\\Objects} & \Centerstack{Human\\Action} & Color& \Centerstack{Spatial\\Relationship} & Scene & \Centerstack{Appearance\\Style} & \Centerstack{Temporal\\Style} & \Centerstack{Overall\\Consistency} \\
\hline

LaVie       & 33.32\% & \textbf{96.80\%} & 86.33\% & 34.09\% & \textbf{52.69\%} & \textbf{23.56\%} & \textbf{25.93\%} & 26.41\%\\ 
ModelScope & 38.98\% & 92.40\% & 81.72\% & 33.68\% & 39.26\% & 23.39\% & 25.37\% & 25.67\% \\ 
VideoCrafter & 25.93\% & 93.00\% & 78.84\% & 36.74\% & 43.36\% & 21.57\% & 25.42\% & 25.21\% \\ 
CogVideo & 18.11\% & 78.20\% & 79.57\% & 18.24\% & 28.24\% & 22.01\% & 7.80\% & 7.70\% \\ \hline
Show-1  & \textbf{45.47}\% & 95.60\% & \textbf{86.35}\% & \textbf{53.5}\% & 47.03\% & 23.06\% & 25.28\% & \textbf{27.46}\% \\
\hline
\end{tabular}
}
\end{table*}
\begin{figure*}[ht]
    \centering
    \includegraphics[width=1\linewidth]{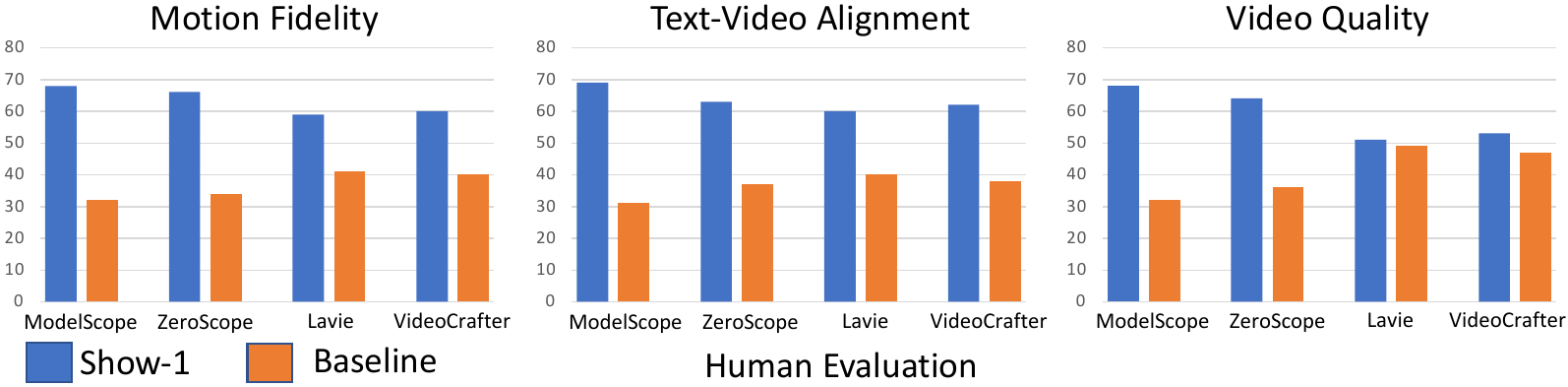}
    \caption{\textbf{Human Evaluations} for ModelScope~\citep{wang2023modelscope}, ZeroScope, VideoCrafter0.9~\citep{chen2023videocrafter1}, LaVie~\citep{wang2023lavie} and our Show-1 model.}
      \label{fig:human}
\end{figure*}

\paragraph{MSR-VTT Experiment.}
The MSR-VTT dataset~\citep{xu2016msr} test subset comprises $2,990$ videos, accompanied by $59,794$ captions. Every video in this set maintains a uniform resolution of $320\times240$. We carry out our evaluations under a zero-shot setting, given that Show-1 has not been trained on the MSR-VTT collection. In this analysis, Show-1 is compared with state-of-the-art models, on performance metrics including FID-vid~\citep{heusel2017gans_nash_equilibrium}, FVD~\citep{unterthiner2018FVD}, and CLIPSIM~\citep{wu2021godivaf}. For FID-vid and FVD assessments, we randomly select 2,048 videos from the MSR-VTT testing division. CLIPSIM evaluations utilize all the captions from this test subset, following the approach ~\citep{singer2022make}. All generated videos consistently uphold a resolution of $256 \times 256$.

\begin{figure*}[t!]
    \centering
    \includegraphics[width=0.83\linewidth]{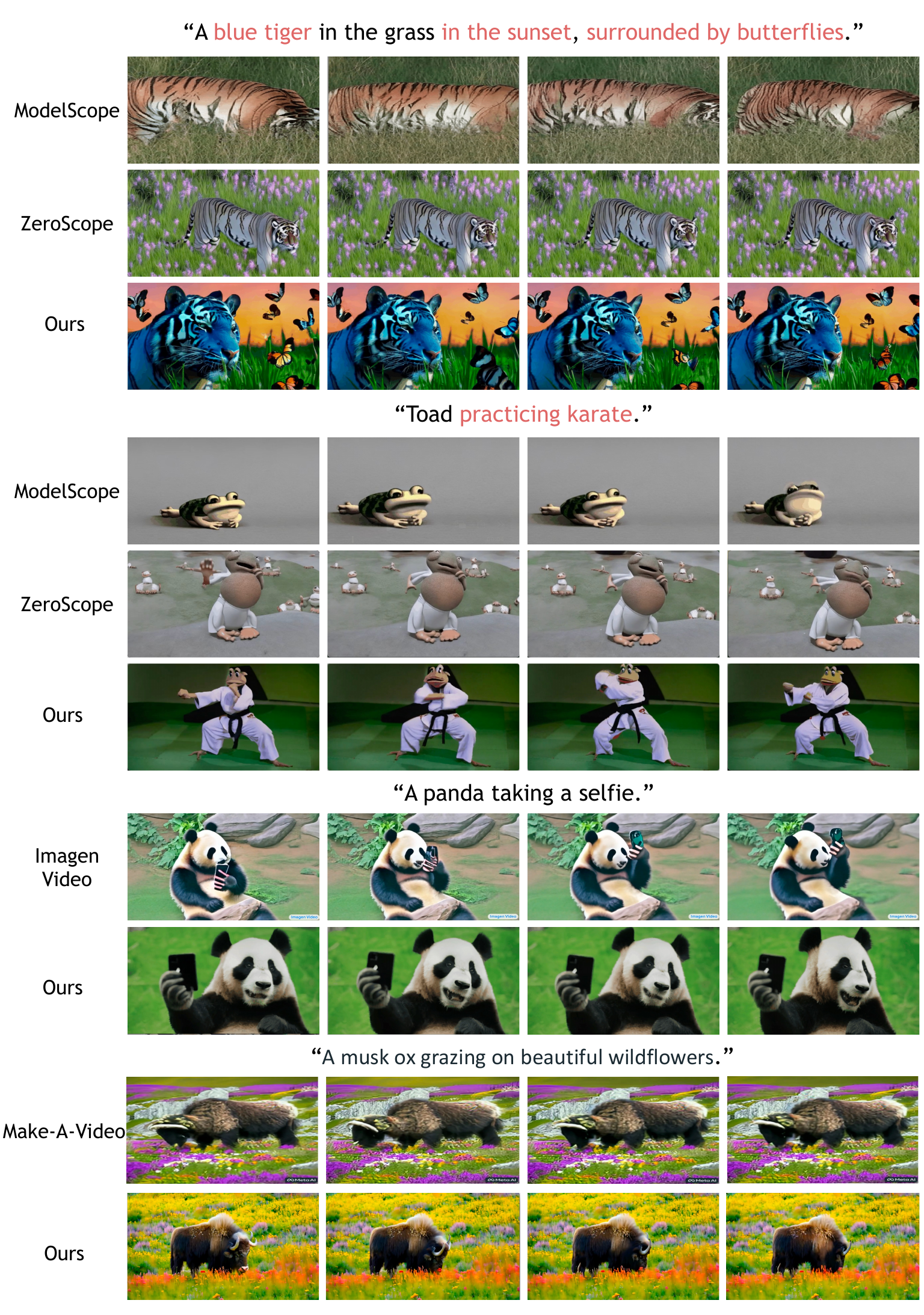}
    \caption{\textbf{Qualitative comparisons with existing video generative models.} Words in red highlight the misalignment between text and video in other open-source approaches (\ie, ModelScope and ZeroScope), whereas our method maintains proper alignment. Videos from closed-source approaches (\ie, Imagen Video and Make-A-Video) are obtained from their websites.}
    \label{fig:compare_1}
\end{figure*}
\begin{figure}[ht!]
    \centering
    \includegraphics[width=1\linewidth]{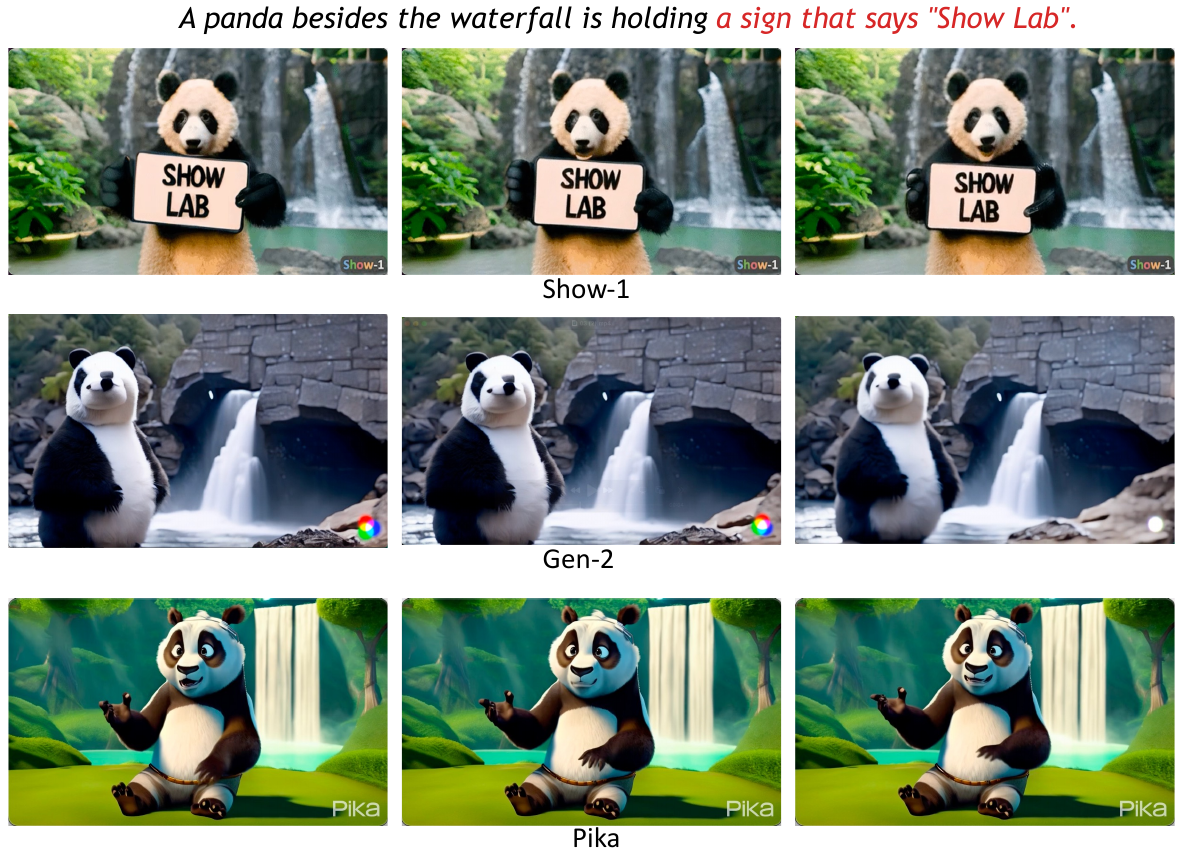}
    \caption{Qualitative comparisons with Gen-2 and Pika. Gen-2 and Pika face challenges in accurately rendering text in videos. Conversely, Show-1 is capable of precise text rendering, indicating superior alignment between text and video.}
    \label{text}
\end{figure}

Tab.~\ref{mrtt} shows that, Show-1  achieves the  second best performance in FID-vid (a score of 12.97) and the best FVD (with a score of 536). This suggests a remarkable visual congruence between our generated videos and the original content. Moreover, our model secures a notable CLIPSIM score of 0.3104, emphasizing the semantic coherence between the generated videos and their corresponding prompts. It is noteworthy that our CLIPSIM score surpasses that of Make-A-Video~\citep{singer2022make}, despite the latter having the benefit of using additional training data beyond WebVid-10M.

\paragraph{VBench Experiment.}
VBench~\citep{vbench} is a benchmark designed for evaluating video generative models by breaking down video generation quality into well-defined dimensions for precise and objective assessment. A Prompt Suite generates videos across various content types for evaluation, while an evaluation method suite offers automated, objective analyses for each dimension. Incorporating human preference annotation ensures VBench's evaluations align with human perceptions, promising valuable insights and open-source availability.

As illustrated in Tab.~\ref{tab:vbench}, out of 16 different evaluation metrics, our approach leads in 10. Notably, these results are obtained by training our Show-1 model on the publicly accessible WebVideo-10M dataset~\citep{bain2021frozen}, marking a significant improvement over VideoCrafter~\citep{chen2023videocrafter1} and LaVie~\citep{wang2023lavie}, which are trained on large-scale, proprietary text-video datasets.

\subsection{Qualitative Results}

\paragraph{Human evaluation.}
We gather an evaluation set comprising 256 complex prompts that encompass camera control, natural scenery, food, animals, people, and imaginative content. The survey is conducted on Amazon Mechanical Turk. Following Make-A-Video \citep{singer2022make}, we assess video quality, the accuracy of text-video alignment and motion fidelity. In evaluating video quality, we present two videos in a random sequence and inquire from annotators which one possesses superior quality. When considering text-video alignment, we display the accompanying text and prompt annotators to determine which video aligns better with the given text, advising them to overlook quality concerns. For motion fidelity, we let annotators determine which video has the most natural notion. As shown in Fig.~\ref{fig:human}, our method achieves the best human preferences on all evaluation parts.

Specifically, our approach exhibits superior text-video alignment and motion fidelity compared to the recently open-sourced ModelScope~\citep{wang2023modelscope}, ZeroScope, VideoCrafter~\citep{chen2023videocrafter1} and LaVie~\citep{wang2023lavie}. Additionally, as depicted in Fig.~\ref{fig:compare_1}, our method matches or even surpasses the visual quality of the current state-of-the-art methods, including Imagen Video and Make-A-Video.  
Furthermore, Show-1 surpasses the commercial products Gen-2 and Pika in terms of text-video alignment, as illustrated in Fig.~\ref{text}.

 \begin{figure*}[ht!]
    \centering
    \includegraphics[width=0.8\linewidth]{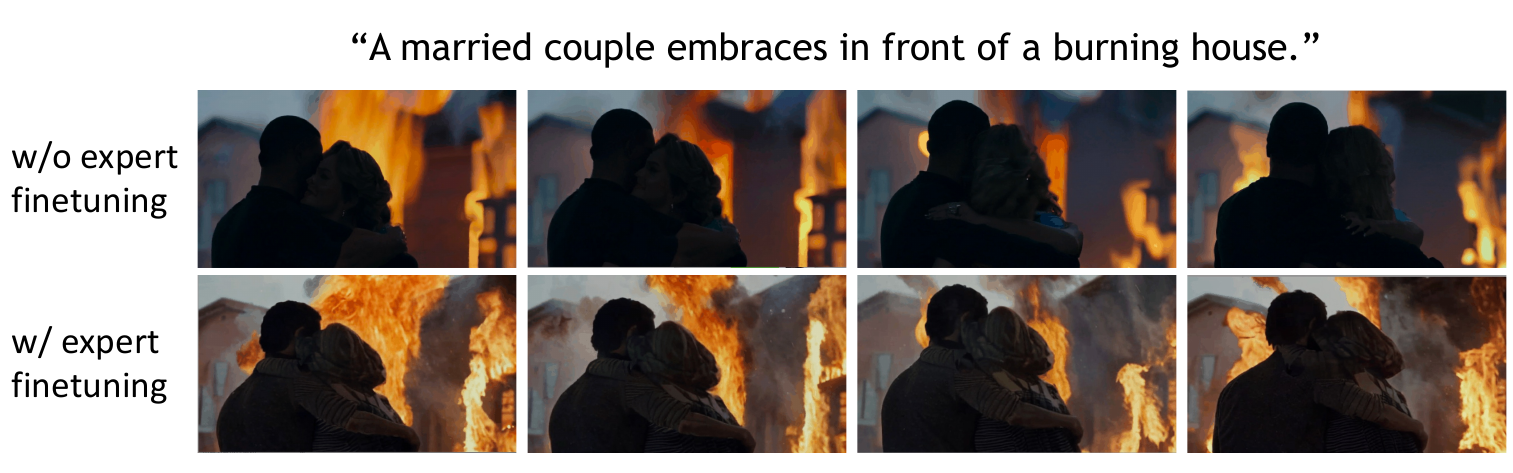}
  
    \caption{\textbf{Effect of expert finetuning.} With expert finetuning, the visual quality is significantly improved.}
    \label{fig:compare_2}
\end{figure*}
\begin{figure}[ht!]
    \centering
    \includegraphics[width=1\linewidth]{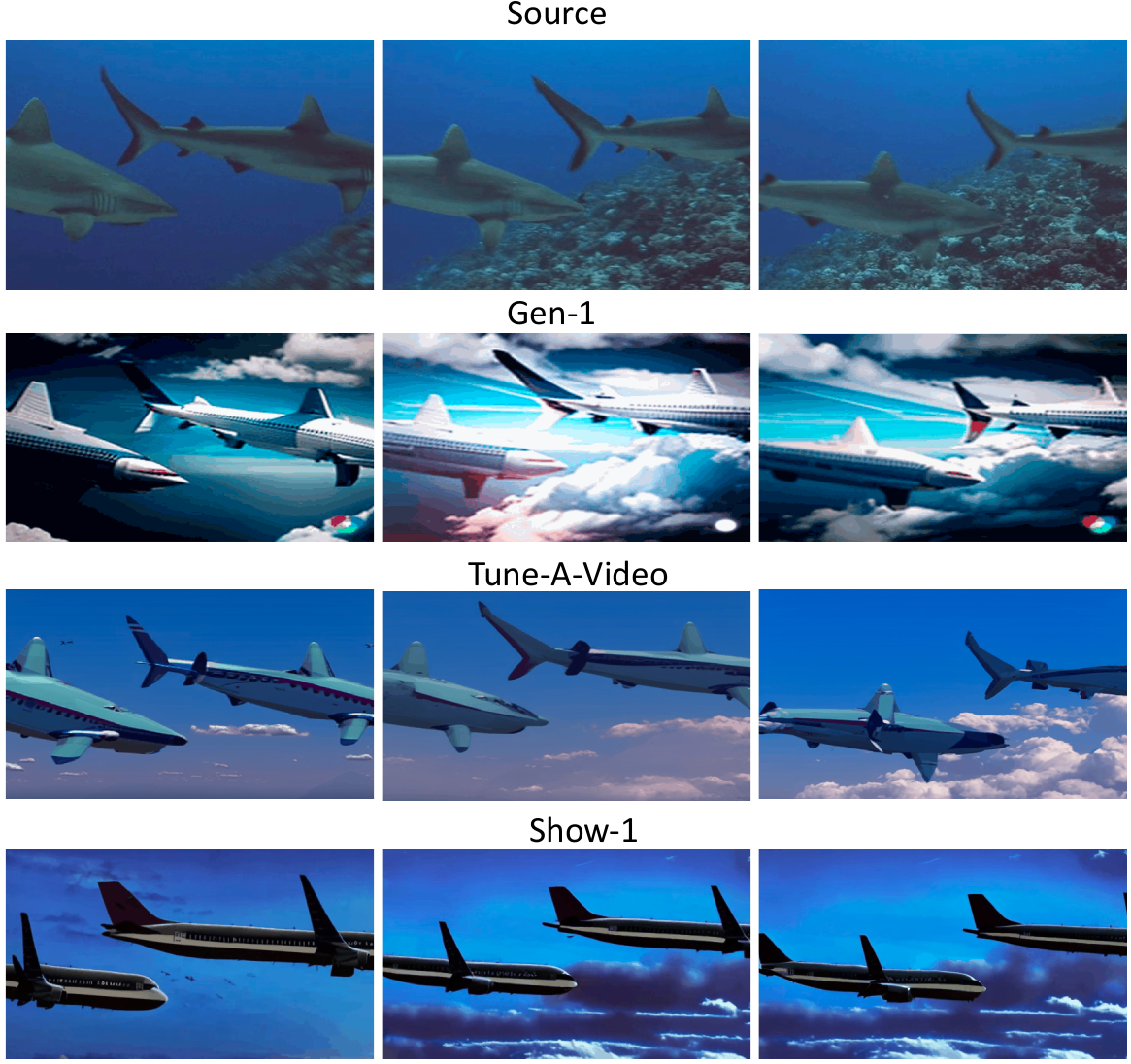}
    \vspace{-3mm}
    \caption{Qualitative comparisons for motion customization. The objective is to utilize the movement from the source video to generate a new video following the prompt \textit{"The planes are flying in the sky."} Other methods often fail to modify the subjects' original shapes in the video, leading to implausible transformations, such as shark-shaped airplanes. In contrast, Show-1 demonstrates superior ability in adapting motion effectively. All results are from ~\cite{jeong2023vmc}.}
    \label{fig:application1}
\end{figure}
\begin{figure}[ht!]
    \centering
    \includegraphics[width=1\linewidth]{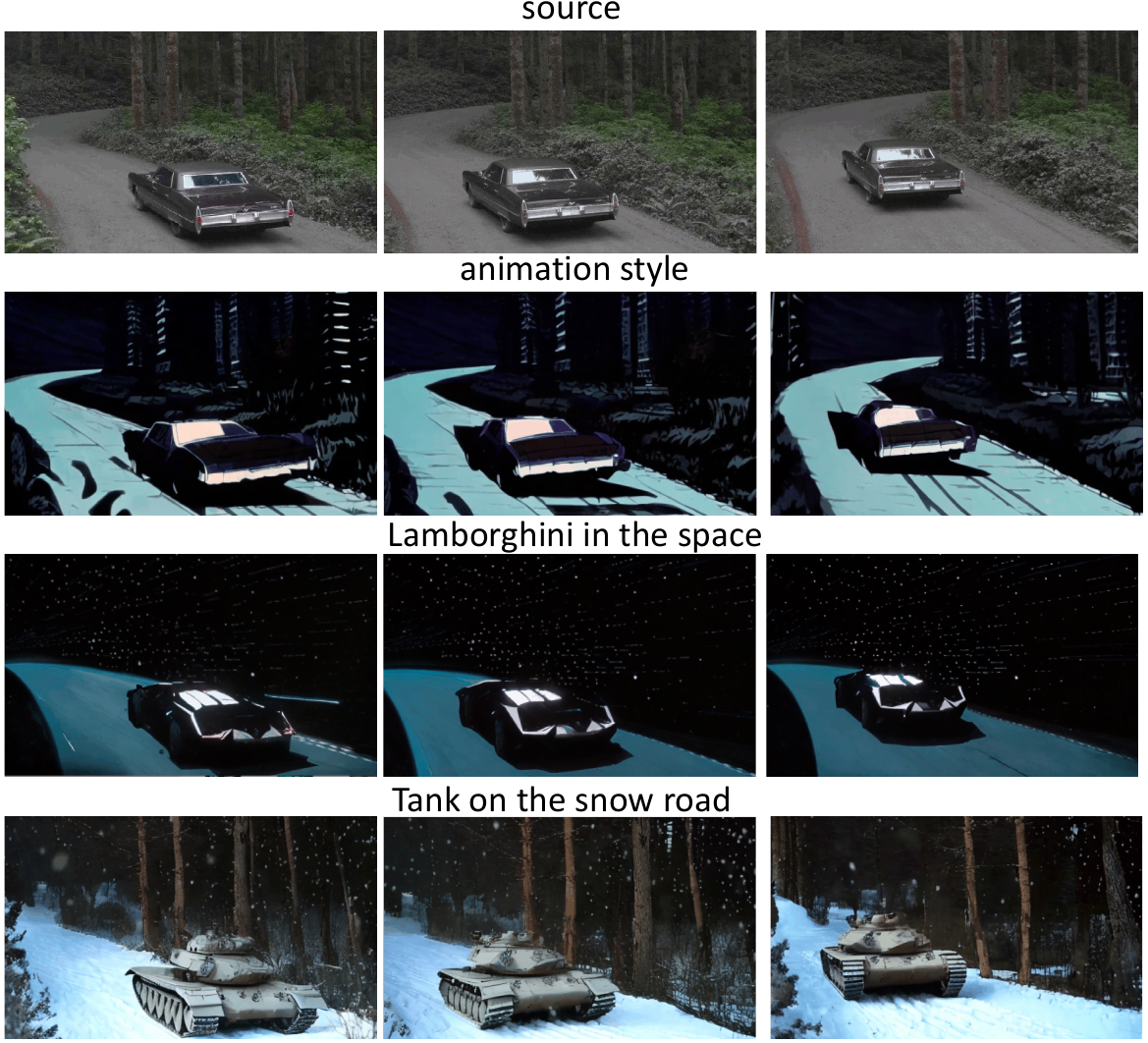}
    \vspace{-3mm}
    \caption{Visualizations for video stylization and editing results of Show-1. All results are from ~\cite{jeong2023vmc}.}
    \label{fig:application2}
\end{figure}

\paragraph{Motion Customization and Video Editing/stylization.}
In Fig.~\ref{fig:application1}, we present visual comparisons of Show-1 with four other methods such as Gen-1~\citep{esser2023structure} and Tune-A-Video~\citep{wu2022tune}. The goal is to harness the motion captured in the original video, where \textit{"sharks are swimming in the sea,"} to create a new video based on the prompt \textit{"The planes are flying in the sky."} Other methods struggle to alter the original form of the subjects in the video, resulting in unrealistic transformations like a shark-shaped airplane. On the contrary, Show-1 excels in customizing motion, managing even complex compositional adjustments successfully, such as depicting sharks or airplanes moving accurately in their respective environments.

As shown in Fig.~\ref{fig:application2}, Show-1 is also capable of delivering impressive results in video stylization and editing that align with the accompanying text. Note that the results are from ~\cite{jeong2023vmc}.


\begin{table*}[ht!]
\centering
\caption{\textbf{Comparisons of different combinations of pixel-based and latent-based VDMs} on keyframes stages and final super-resolution stage in terms of text-video similarity, memory usage during inference, UCF-101 FVD and  human evaluations of motion fidelity. The same T5 text encoder  is employed across both pixel-based and latent-based VDMs in the keyframes stage. Each model is initialized with the image model weights pre-trained on the LAION dataset and has the same number of parameters, maintaining consistency for fair comparisons. $f$ indicates the latent compression ratio.} 
\label{ablation1}
 \centering
	\resizebox{0.85\textwidth}{!}{
\begin{tabular}{l c|c cccc}
\hline
  \Centerstack{Keyframes\\Stage} &  \Centerstack{Final Super-Res.\\ Stage} & \Centerstack{ CLIP\\SIM} &  \Centerstack{Max \\ Memory} &  \Centerstack{UCF-101 \\ FVD$\downarrow$} &  \Centerstack{Text-Video \\Alignment}  &  \Centerstack{Motion\\ Fidelity}   \\
\hline
-- & pixel & -- & 72GB &-- &-- & --\\
$64\times 40/$ pixel & latent &\textbf{0.3096}&  15GB &\textbf{383} & \textbf{36\%} & \textbf{23\%}  \\
\hline
$64\times 40/$ latent $f=8$ & latent & 0.2441  &  15GB & 584 & 1\% & 2\% \\
$64\times 40/$ latent $f=4$  & latent & 0.2524  &  15GB & 552 & 1\% & 5\% \\
$64\times 40/$ latent $f=2$& latent & 0.2742  &  15GB & 465 & 2\% & 4\% \\
\hline

$256\times 160/$ pixel  & pixel &0.2784 &  48GB & 462 &3\% & 6\% \\
\hline
$256\times 160/$ latent $f=8$ & latent &0.2874 &  15GB & 416 &11\% & 15\% \\
$256\times 160/$ latent $f=4$ & latent &0.2897 &  15GB & 403 &16\% & 11\% \\
$256\times 160/$ latent $f=2$ & latent &0.2834 &  26GB & 429 &8\% & 9\% \\

\hline
$512\times 320/$ latent $f=8$ & latent &0.2793 & 15GB &487 & 7\% &10\%\\
$512\times 320/$ latent  $f=4$& latent &0.2879 & 26GB &426 & 9\% &9\%\\
$512\times 320/$ latent $f=2$& latent &0.2767 & 48GB &451 & 6\% &6\%\\

\hline
\end{tabular}}
\end{table*}
 \begin{table*}[ht!]
\centering
\caption{\textbf{Comparisons of parameters and speed between Make-A-Video and our method.} The numbers are reported in the format of \textit{Make-A-Video~\citep{singer2022make} / ours}}
\label{table:speed}
\resizebox{0.82\textwidth}{!}{
\begin{tabular}{lcccccc}

\hline
Stage & Prior & Keyframes& Temp. Interp. & Super1 &Final Super & Total \\
\hline
Step & 64/- & 100/ 75 & 50/ 75 & 50/ 50 & 50/ 40 & - \\
Para.& 1.3B/ - & 3.1B/ 1.7B & 3.1B/ 1.7B & 1.4B/ 0.8B & 0.7B/ 1.8B & 9.6B/  6B \\
Time & 3s/ -- & 58s/ 30s & 62s/ 60s & 70s/ 65s & 63s/ 23s & 256s/ 178s\\
Memory &7GB/ --  & 18GB/ 11GB & 14GB/ 10GB &52GB/ 14GB & 54GB/ 15GB &--/-- \\
FVD & -- & 569   & 542 &   474  & 383 &--\\
\hline
\end{tabular}}
\end{table*}

\begin{table}[t]
\centering
\caption{\textbf{Ablation study of our final super-resolution module on UCF-101.}}
\label{tab:ablation2}

\centering
\resizebox{0.41\textwidth}{!}{
\begin{tabular}{lcc}
\hline
Methods & FVD($\downarrow$) & IS($\uparrow$) \\
\hline
Sdx4 with temporal & 459 & 32.98\\
\hline
Expert translation & &  \\
\quad change input & 423& 33.83 \\
\quad + expert finetuning & 383& 35.67\\
\hline
\end{tabular}}
\vspace{-2.5mm}
\end{table}

\subsection{Ablation Studies}

\paragraph{Decide which stage should use pixel or latent, whether to generate high resolution or low resolution.}
The initial step involves determining the resolution and the VDM employed for the keyframe stages. As illustrated in Tab.~\ref{ablation1}, utilizing a pixel-based VDM with a low resolution of $64 \times 40$ outperforms the corresponding latent model ($f=8$) at the same resolution. This suggests the difficulty for a small latent space (e.g., $8 \times 5$ for videos of $64 \times 40$ resolution) to capture the comprehensive and necessary visual semantic details as outlined by the text prompt. Additionally, the $64 \times 40$ pixel VDM also outshines the $256 \times 160$ latent-based VDM in performance, and when the resolution is increased to $512 \times 320$, the results diminish, indicating that the latent model may focus more on spatial appearance at higher resolutions, possibly neglecting alignment with the text.Meanwhile, at a resolution of $64 \times 40$, the text-video alignment significantly decreases with larger $f$ values. At a resolution of $256 \times 160$ and $512x320$, all values of $f (0, 2, 4, 8)$ result in worse text-video alignment and efficiency compared to $64 \times40$ with $f=0$. In conclusion, these findings indicate that starting with very low-resolution keyframes using pixel-based VDM($f=0$) yields the best alignment between video and text, along with motion quality. Given that subsequent stages also work with low-resolution video, pixel-based VDM is chosen for these phases as well. However, due to the significantly higher computational cost of pixel models at high resolutions, as shown in Tab.~\ref{ablation1}, we opt for the latent VDM for our final super-resolution module.

\paragraph{Impact of expert translation of latent-based VDM as final super-resolution module.} We present ablations with and without the incorporation of expert translation. Detailed in Section \ref{section3.6}, "expert translation" involves two key changes: modifying the UNet input and implementing expert fine-tuning, which entails training the latent-based VDMs over timesteps 0-900 out of a maximum of 1000. According to Tab.~\ref{tab:ablation2}, models enhanced with expert translation yield videos of higher quality compared to the standard SDx4 model equipped with the temporal layers. Furthermore, as depicted in Fig.~\ref{fig:compare_2}, the visuals demonstrate that our expert fine-tuning approach results in reduced artifacts and captures more complex details.

\paragraph{Inference Speed.} 
Although hierarchical structures require more inference time compared to single-stage models, their outcomes are significantly superior, as evidenced by advanced generation methods like~\citep{ho2022imagen,singer2022make,blattmann2023align}. These SOTA methods all employ hierarchical frameworks, including keyframe generation, temporal interpolation, and superresolution, for video creation. We replicated the Make-A-Video model by precisely matching its parameters and network architecture for inference time  and parameters comparisons. As shown in  Tab.~\ref{table:speed}, the results show that our method is faster and more memory efficient than the previous SOTA method, Make-A-Video.

\section{Conclusion}

We introduce Show-1, a novel model that marries the strengths of pixel and latent based
VDMS. Our approach employs pixel-based VDMs for initial video generation, ensuring precise
text-video alignment and motion portrayal, and then uses latent-based VDMs for super-resolution,
transitioning from a lower to a higher resolution efficiently. This combined strategy offers high-quality text-to-video outputs while optimizing computational costs.

\section*{Ackmowledgement}

This research is supported by the Ministry of Education, Singapore, under the Academic Research Fund Tier 1 (FY2023). The computational work for this article was partially performed on resources of the National Supercomputing Centre, Singapore.

\bibliographystyle{spbasic}      
\bibliography{ref}   

\end{document}